\documentclass[letterpaper, 10 pt, conference]{ieeeconf} 
\IEEEoverridecommandlockouts  
\usepackage[noadjust]{cite}

\usepackage{amsmath,amssymb,amsfonts}
\usepackage{algorithmic}
\usepackage{graphicx}
\usepackage{textcomp}
\usepackage{xcolor}
\usepackage{lipsum}
\usepackage{booktabs}
\usepackage{multirow}
\usepackage{colortbl}
\usepackage{caption}
\usepackage{multicol}
\usepackage{pifont}
\usepackage{hyperref}

\newcommand{\xmark}{\text{\color{red}\ding{55}}}

\def\model{ReMEmbR}
\def\dataset{NaVQA}

\def\BibTeX{{\rm B\kern-.05em{\sc i\kern-.025em b}\kern-.08em
    T\kern-.1667em\lower.7ex\hbox{E}\kern-.125emX}}
\begin{document}

\title{ReMEmbR: Building and Reasoning Over \\Long-Horizon Spatio-Temporal Memory for Robot Navigation}


\author{Abrar Anwar$^{1,2}$, John Welsh$^{1}$,  Joydeep Biswas$^{1,3}$, 
Soha Pouya$^{1}$, Yan Chang$^{1}$ 
\thanks{
$^{1}$ NVIDIA, $^{2}$ University of Southern California, $^{3}$ University of Texas at Austin. This work was done while Abrar Anwar was an intern at NVIDIA}
\thanks{
    Contact:
    \tt{abrar.anwar@usc.edu, \{jwelsh, jbiswas, spouya, yachang\}@nvidia.com}
}
}

\maketitle

\begin{abstract}
Navigating and understanding complex environments over extended periods of time is a significant challenge for robots.
People interacting with the robot may want to ask questions like where something happened, when it occurred, or how long ago it took place, which would require the robot to reason over a long history of their deployment.
To address this problem, we introduce a Retrieval-augmented Memory for Embodied Robots, or {\model}, a system designed for long-horizon video question answering for robot navigation.
To evaluate {\model}, we introduce the {\dataset} dataset where we annotate spatial, temporal, and descriptive questions to long-horizon robot navigation videos.
{\model} employs a structured approach involving a memory building and a querying phase, leveraging temporal information, spatial information, and images to efficiently handle continuously growing robot histories.
Our experiments demonstrate that {\model} outperforms LLM and VLM baselines, allowing {\model} to achieve effective long-horizon reasoning with low latency.
Additionally, we deploy {\model} on a robot and show that our approach can handle diverse queries. 
The dataset, code, videos, and other material can be found at the following link: \href{https://nvidia-ai-iot.github.io/remembr}{https://nvidia-ai-iot.github.io/remembr}
 \end{abstract}


\section{Introduction}
Robots are increasingly being deployed in a wide variety of environments, including buildings, warehouses, and outdoor settings.
During their deployments, robots perceive a range of objects, dynamic events, and phenomena that are challenging to encapsulate within conventional representations like metric or semantic maps. 
Additionally, these robots exist for long periods of time, typically on the magnitude of hours, but there is currently no way to query the robot on what it has seen over this long period of time. 
In this work, we address the challenge of efficiently building this long-horizon memory for robot navigation and responding to questions by framing it as a long-horizon video question-answering task.
Our system enables robots to respond to free-form questions and to perform actions based on what they have observed.

\begin{figure}
\centering
\includegraphics[width=.95\columnwidth]{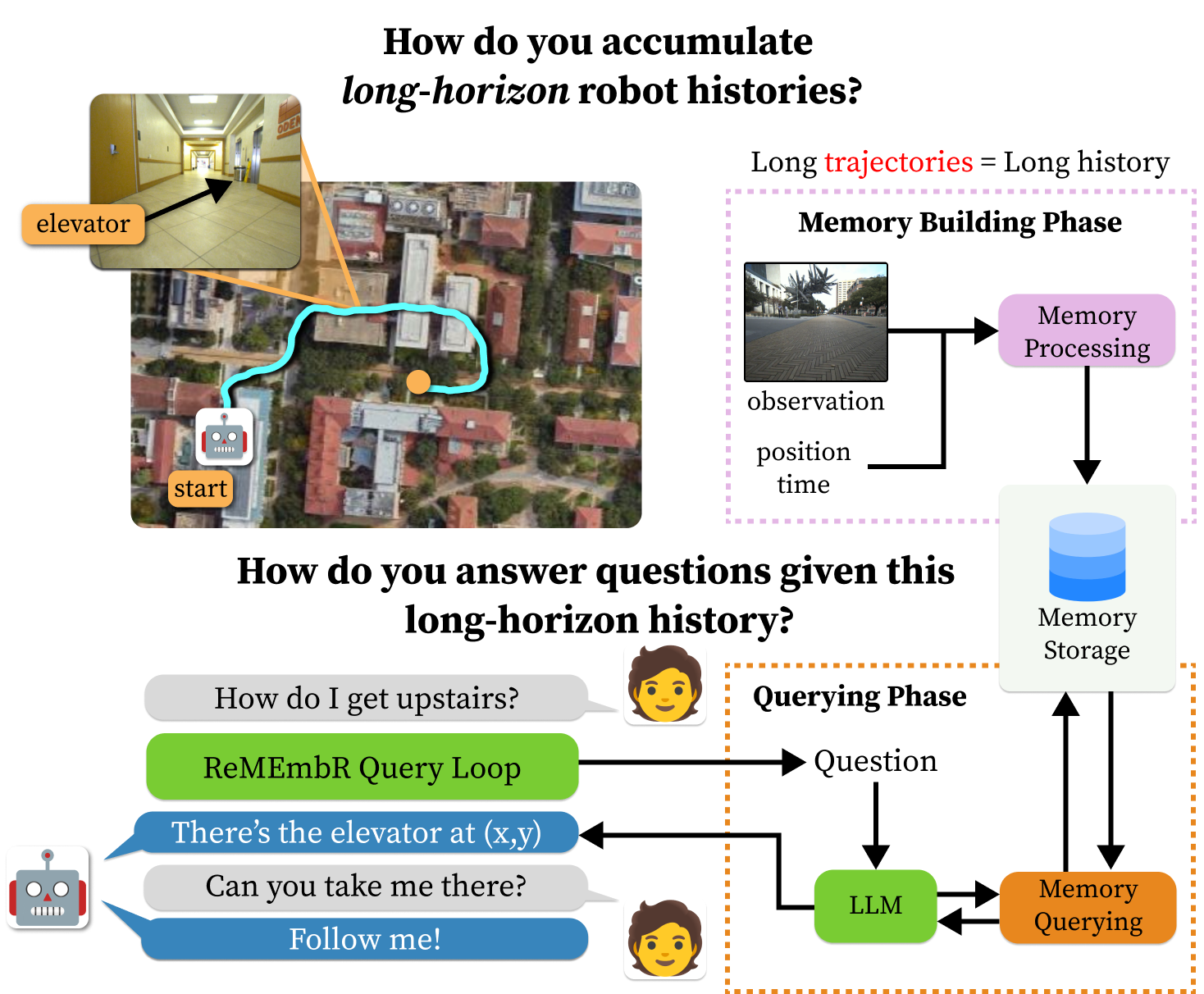}
\caption[]{
Robots continuously operate for long periods of time, where they gather long histories. 
In this work, we investigate how to aggregate these robot histories over time efficiently, and how to utilize that memory representation for answering spatio-temporal questions and generating navigational goals.
}
\label{fig:teaser}
\end{figure}

Existing approaches to spatio-temporal video memory in robotics are constrained by their capacity to handle only short durations, typically limited to 1-2 minutes~\cite{chen2023open,robovqa2023arxiv}. 
As the time span increases, the inference time memory requirements grow for transformer-based methods, rendering them impractical for processing arbitrarily long videos
Concurrent work~\cite{chiang2024mobility} has focused on leveraging extremely large context windows of large language models to answer questions given a long robot history; however, this is not a scalable solution.
No matter the length of the context window, unbounded length histories will not fit in fixed context sizes.
In this work, we propose a Retrieval-augmented Memory for Embodied Robots, or {\model}, which uses a retrieval-based LLM-agent capable of querying memory across arbitrary lengths by formulating text-based, spatial, and temporal queries.
As shown in Figure~\ref{fig:teaser}, {\model} consists of a memory-building phase and a querying phase.


Episodic memory in robotics has been framed mostly as a question-answering task, where systems are evaluated based on their ability to answer questions from a given video~\cite{das2018embodied, grauman2022ego4d}. 
While useful for assessing QA capabilities, the text-based outputs of these systems may fall short of providing actionable information for a navigation robot.

For example, a question like ``Where did you see my phone?" might yield a response such as ``I saw it on the coffee table." 
While informative, this answer does not translate into actionable data for the robot. 
Our work, therefore, also incorporates reasoning over explicit spatial (e.g., xy positions) and temporal (e.g., ``10 minutes ago") information.

To evaluate our system, we construct the Navigation Video Question Answering dataset {\dataset} where methods must output position, temporal information, or free-form text. 
Our dataset consists of $210$ questions sampled from subsets of 7 long-horizon navigation videos.
This dataset is intended to foster further research in long-horizon memory building and reasoning for navigation robots.

In particular, we

\begin{itemize}

    \item design the {\dataset} dataset for evaluating 1) whether a robot had seen events or objects over the course of its deployment, 2) when it saw certain events or objects, 3) where they happened, and 4) how to reason about these spatio-temporal aspects to answer questions;
    \item introduce {\model}, a retrieval-augmented LLM-agent capable of forming function calls to retrieve relevant memories and answer questions based on a real-time memory-building process;
    \item provide qualitative results on a real-world deployment of {\model} on a robot, testing whether {\model} is able to reason over its long-horizon deployment.

\end{itemize}

\section{Related Work}

\textbf{Embodied question answering.}
Embodied Question Answering (EQA)~\cite{das2018embodied, das2018neural, majumdar2024openeqa, ren2024explore, thomason2018shifting, wijmans2019embodied} is an extension of video question answering to egocentric, and possibly interactive, environments, requiring agents to navigate and gather information to answer questions. 
Most similar to the question answering ability of our work is OpenEQA~\cite{majumdar2024openeqa}, which answers questions about what a robot has seen.
However, their questions consider only a short 30-second memory.
This formulation falls short when applied to robotics scenarios that involve extended time horizons and continuous interaction with the environment.
In our work, we focus on answering questions and generating navigational goals on longer lengths of history and leverage robot-centric data such as position and time.

\textbf{Language and navigation.}
Classical navigation typically uses metric maps and does not focus on navigating to semantic goals. 
Most recent work in vision-and-language navigation~\cite{anderson2018vision, thomason2020vision,zhou2024navgpt, krantz2023iterative, gan2022vision}, object-goal navigation~\cite{chaplot2020object, gadre2023cows,majumdar2022zson,ramrakhya2022habitat}, and various forms of language-guided navigation~\cite{shah2023lm,shah2023navigation,dorbala2022clip} focus on navigating in unseen spaces.
These works focus more on exploration; however, robots typically are deployed for extended periods of time in the same area.
Forms of memory such as scene graphs~\cite{wald2020learning,li2022embodied}, topological memory~\cite{eysenbach2019search,savinov2018semi}, or queryable map representations~\cite{fang2019scene,chen2023open,shafiullah2022clip} may also allow for semantic goal generation, but may fall short in answering questions about a robot's experience over time about dynamic, non-static objects.
As such, our work uses a robot's video to capture these details over a robot's deployment.

MobilityVLA~\cite{chiang2024mobility} is a concurrent work where a long-horizon robot video tour is given to the 1M length context window of a Gemini LLM from which the robot must generate a topological goal.
In this work, we solve a more general problem of answering spatial, temporal, and descriptive questions while also generating metric navigation goals.
Additionally, simply increasing the context window length is not scalable to unbounded history lengths.
Using retrieval-based methods, our approach can scale better to long histories.

\textbf{Large language models and robotics.}
Recent years have seen advancements in large language models (LLMs) and vision-and-language models (VLMs), significantly expanding their capabilities across various tasks~\cite{touvron2023llama,achiam2023gpt}.
Prompting techniques such as chain-of-thought~\cite{wei2022chain} and others~\cite{yao2024tree,zhou2022least} has further enhanced LLMs' problem-solving abilities, enabling more complex reasoning.
Retrieval-augmented generation~\cite{gao2023retrieval,asai2023self} and LLM-agents~\cite{shinn2024reflexion,yao2022react,park2023generative, schick2024toolformer} allow the LLM to leverage external information to provide further context to the LLM.
In robotics, past work have used the reasoning ability of LLMs for task planning~\cite{brohan2023can, huang2022inner, singh2024twostep, liu2023llm}, generating plans as code~\cite{singh2023progprompt,hu2024deploying,liang2023code}, or to generate navigational goals~\cite{chiang2024mobility, shah2023lm}.
Rather than focusing on planning, our work focuses open horizon perception, and builds an LLM-agent to enable scalable multi-step reasoning over long-horizion robot histories.




\begin{figure*}[ht]
\centering
\vspace{5px}
\includegraphics[width=1\linewidth]{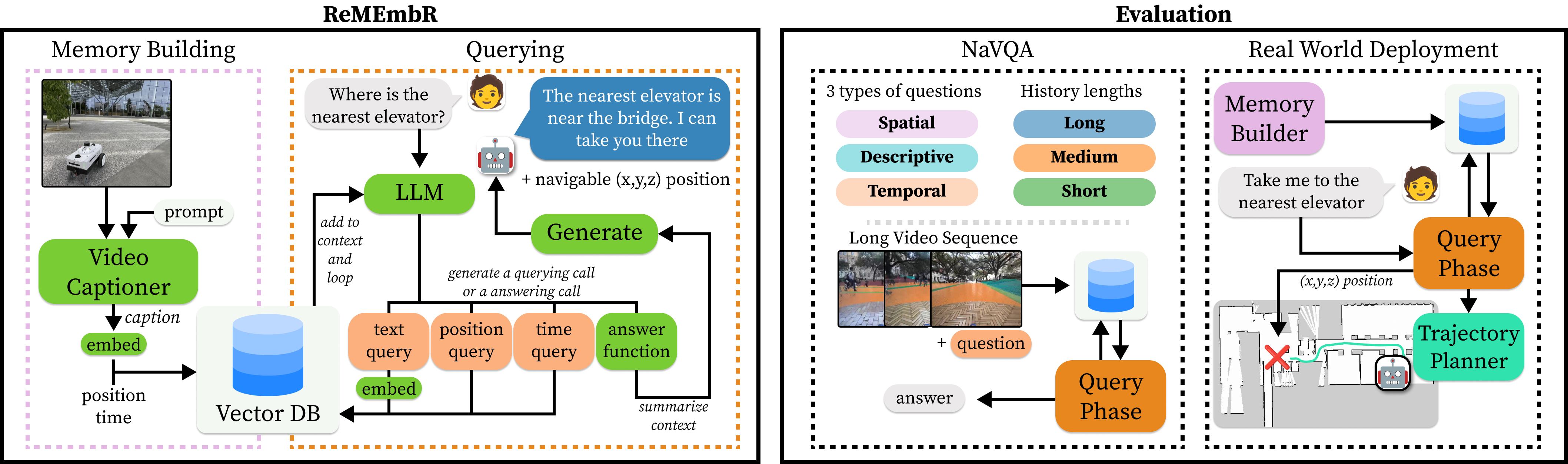}
\caption[]{(Left) We design {\model} with a memory building phase and a querying phase. The memory building phase runs a VILA~\cite{lin2024vila} video captioning model, embeds the caption, then stores the caption embedding, position, and time vectors into a vector database. Then, when a user asks a question, a vector database querying loop starts with an LLM. (Right) Then, we evaluate {\model} on the {\dataset} dataset which we construct. {\dataset} consists of three types of questions as shown above. Then we deploy {\model} on a robot.}
\label{fig:overview}
\end{figure*}

\section{Problem Formulation}

We formulate our problem as a variation of a long-horizon video question answering task for robots.
Unlike standard video question answering, robots are deployed for $K$ minutes and actively accumulate a history $H_{1:K}$ from various sensors.
Due to their continuous deployment, the size of the history is monotonically increasing over time.
Thus our work focuses on two problems: efficiently building a representation this long history $H_{1:K}$ over time and then querying the representation to answer questions and generate navigational goals.


To efficiently build a memory, we consider a history of images $H_I$, positions $H_P$, and timestamps $H_T$.
We assume that the robot has localization capabilities, such as using LIDAR-based localization, GPS, or odometery information to provide metric coordinates.
After a memory representation is built, a user asks the robot a question $Q$ about spatial, temporal, or descriptive information which the robot may have seen.
Specifically, our goal is to predict an answer $A$ given the history and a question $p(A|Q,H)$.

\textbf{Questions.}
Robots need to localize information in their histories; however, we focus on making this information actionable.
For \textbf{spatial} questions such as ``Where is closest bathroom?", the robot must reason about all the bathrooms and signs for bathrooms it has seen.
Then, the system must provide the specific (x,y) location to go to the closest bathroom.
By formulating spatial questions with coordinates, robots can act on this information to navigate to these goals.

Users may also want to query how long ago an event had occurred or understand how long a robot has done a task. 
Thus, we define two types of \textbf{temporal} questions: point-in-time questions and duration questions.
Point-in-time questions such as ``When did you see the boxes fall?" with the answer ``15 minutes ago" refer to a specific point-in-time relative to the present.
Duration questions focus on the length of an activity such as ``How long were you inside the building for?" with the answer ``10 minutes". 
These temporal questions allow robots to retrospectively consider their previous actions.

Lastly, \textbf{descriptive} questions ask about the environment, activities the robot may have seen, or the robot's state in the past.
This general category can be yes or no questions such as ``Was the sidewalk busy today?" or be more descriptive like ``What side of the street are you driving on?"
These descriptive questions ensure that our robots can effectively remember pertinent details that users ask for.


To capture these questions, we build the {\dataset} dataset. 
We then design {\model} as a step towards solving this task.

\section{\model}

Since robots are embodied and continually persist in the environment, we decompose the task into two distinct phases: memory building and querying. 

The computation of $p(A|Q,H_{1:K})$ is often difficult, as long histories are computationally expensive for Transformer-based models or can lead to forgetting in state-space models such as LSTMs. 
We note that for a given question, a large history is often not required to provide a correct answer.
Instead, only a subset of the history $R \subseteq H_{1:K}$ is needed.

Therefore, we can compute the answer given an optimal history subset $R^* \subseteq H_{1:K}$.
In practice, we cannot compute $R^*$ and must sample an $R$ such that it contains the same information as $R^*$.
To do so, we build a memory representation $V$ that is sampled using $F: V \rightarrow R$, where $F(V) = \{h | h \in H_{1:K}\}.$
We decompose the problem as follows:
\begin{equation}
    p(A|H_{1:K}, Q) = p(A|R^*, Q) \approx p(A|{R}, Q), 
\end{equation}
$\text{ s.t. } {R} \sim F(V)$. 
Then, our goal is to estimate $R^*$ such that the answers derived from $R$ and $H$ are consistent.
To do so, we must minimize the size of $R$ while ensuring that the answer can be predicted from both the history $H$ and the subset $R$:
\begin{equation}
    \begin{gathered}
        R^* = \underset{R}{\mathrm{argmin}} \; |{R}| \\ 
        \text{ s.t. } \underset{A}{\mathrm{argmax}} \; p(A|R, Q) = \underset{{A'}}{\mathrm{argmax}} \; p(A'|H,Q)
    \end{gathered}
\end{equation}



Using a memory representation $V$ and a sampling strategy $F$ makes the computation more tractable given a long history. 
Next, we detail how {\model} aggregates the memory representation $V$ during a memory building phase and how it samples $R \sim F(V)$ during a querying phase.

\textbf{Memory Building.}
As robots aggregate information over time, we define the queryable memory representation $V$ as a vector database.
Vector databases are commonly used to store millions of vector embeddings and search efficiently through them using quantized approximate nearest neighbor methods.
Since these databases are efficient in search, we use a vector database to store time, position, and visual representations.

Robots perceive static objects, scenes, and dynamic events, over the course of their deployments. 
We would like to note that the memory representation $V$ must be constructed without knowing the question $Q$ in advance, and thus must be general enough for any potential question.
As the robot is moving in real-time, we aggregate $t$ seconds of image frames $H_{I_{i:i+t}}$ to compute an embedding representation for that segment of memory.
We use video captioning using VILA~\cite{lin2024vila} over each consecutive $t$-second segment, which generates a caption for each temporal segment $L_{i:i+t}$.
These captions capture low-level details of what the robot sees over time, which we then embed using a text embedding function $E$. 
We use the mxbai-embed-large-v1~\cite{mxbai} embedding model to embed the captions.
Over time, the robot adds the vector representation of the text captions, the position, and the timestamps $E(L_{I_{i:i+t}}), H_{P_{i:i+t}}, H_{P_{i:i+t}}$ into the vector database $V$.

\textbf{Querying}
With the vector database $V$ in place, the querying phase can begin.
To gather a history subset $R$, we use an LLM-agent as the sampling function $F$ to sample the database $V$.

The LLM-agent acts as a state machine that iteratively calls the LLM as shown in Figure~\ref{fig:overview}. 
Our approach begins with a retrieval node which queries the vector database in three different ways, using position, timestamp, or text embeddings.
The LLM considers the current set of memories $R_{0:i}$ and the question $Q$ to generate a function call $f$ and a query $q$ which retrieves $m$ memories. 
Each memory contains position, time, and caption information to be used as further context.
These $m$ retrieved memories are then added into $R$:
$$R_{i:i+m} = f(q) \text{, where } q=LLM(R_{0:i}, Q).$$

We define three functions which the LLM could call:

\begin{itemize}
    \item \textbf{Text retrieval}: $f_l(\text{object}) \longrightarrow m$ memories  
    \item \textbf{Position retrieval}: $f_p({(x,y,z)}) \longrightarrow m$ memories  
    \item \textbf{Time retrieval}: $f_t(\text{"HH:MM:SS"}) \longrightarrow m$ memories  

\end{itemize}

At each iteration, the LLM can formulate up to $k$ queries that may help it answer the question.
Once $k \times m$ memories are retrieved, the LLM assesses whether the question can be answered with the updated context.
If the question is not answerable, the LLM uses the current context and executes the querying phase again retrieve new memories.
If the question is answerable, the LLM summarizes any relevant information, and then generates an answer given the entire based on all the retrieved memories.
The output is formatted as a JSON with keys for text, position, time, or duration answers. 
This structured output ensures simple evaluation on {\dataset} and makes it easy to generate goals for a robot deployment.

\begin{figure}[t]
\centering
\vspace{5px}
\includegraphics[width=1\columnwidth]{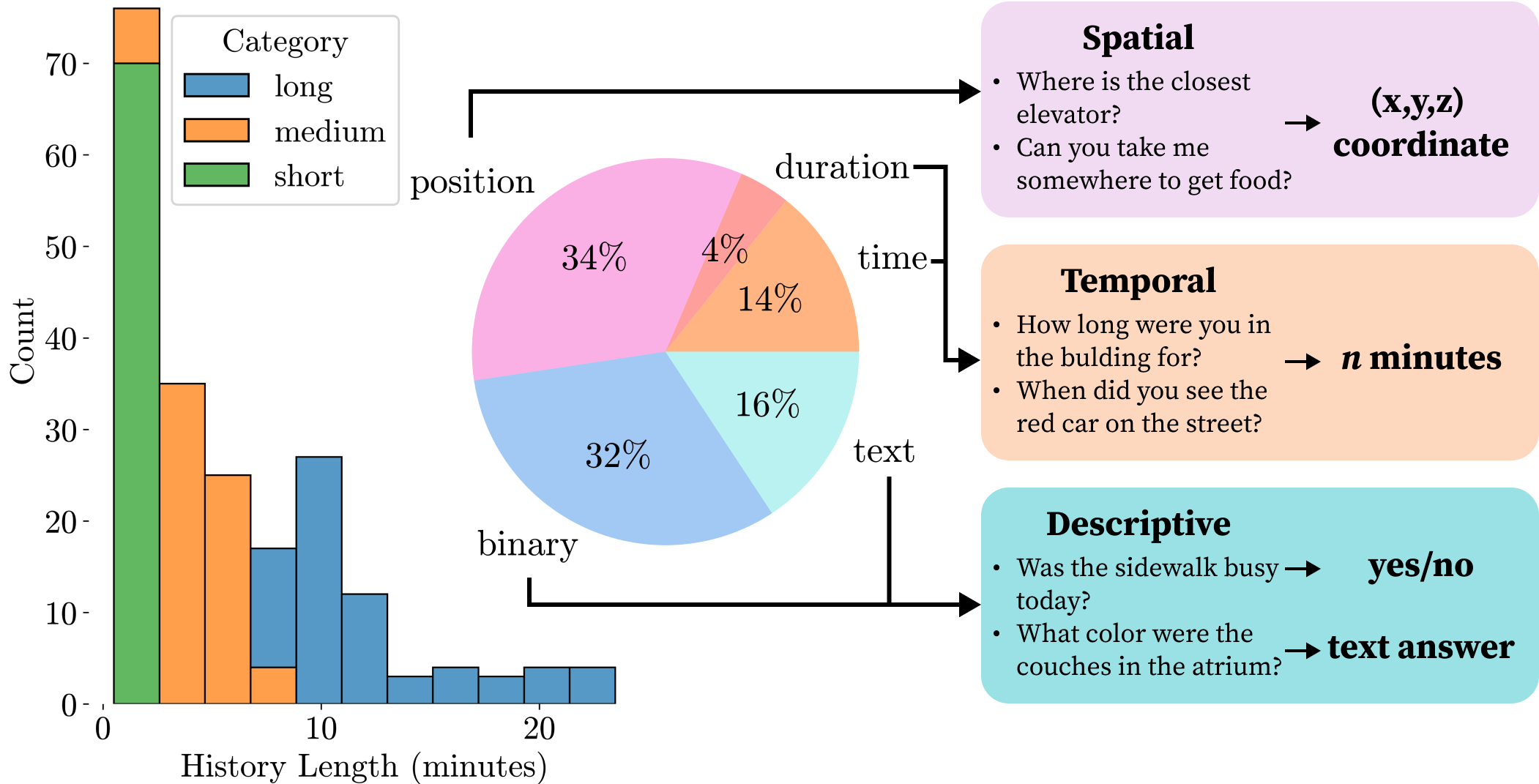}
\caption[]{We introduce the {\dataset} dataset, which is composed of $210$ examples across three different time ranges up to 20 minutes in length. The dataset consists of spatial, temporal, and descriptive questions, each of which has different types of outputs as shown above.}
\label{fig:dataset}
\end{figure}

\section{Dataset}

We introduce the {\dataset} dataset, a long-horizon navigation video question answering dataset built on top of the CODa robot navigation dataset \cite{zhang2024towards}. 
As described in the previous section, this dataset is annotated with spatial, temporal, and descriptive questions and answers.
We use these questions to evaluate models' ability to handle robot-centric long-horizon reasoning.
We are excited for the robot learning community to leverage this robot-centric QA dataset to improve the long-horizon reasoning capability of robots.

\textbf{CODa Dataset.}
The CODa dataset is a large urban navigation dataset consisting of long-horizon sequences in indoor and outdoor settings on a university campus.
The dataset was collected using a Clearpath Husky robot \cite{ClearpathHusky}, where the robot navigated during the morning, afternoon, and evening. 
This data is also realistic to what outdoor robots may encounter, with sunny, cloudy, low-light, and rainy sequences. 
Though the dataset provides various sensor information such as LIDAR, GPS, LIDAR, and multiple cameras, we consider only the GPS coordinate and a front-facing camera. 
We select 7 of the 23 sequences in the CODa dataset for building the {\dataset} dataset.
Each sequence ranged in length from 15 to 30 minutes.

\textbf{Data Annotation.}
We are interested in how varying the length of a robot trajectory may impact the question answering ability of a system.
In our work, we subsample the 7 sequences into three length-based categories: less than 2 minutes \textit{(short)}, between 2 and 7 minutes \textit{(medium)}, and longer than 7 minutes \textit{(long)} segments.
For each sequence, we subsample $10$ segments of each length category, which we then provide to annotators to design questions and answers for.
This process leads to $30$ questions per sequence, for a total of $210$ total questions.
As these videos are long and require an understanding of robot perception, we recruited $5$ robot experts to annotate spatial, temporal, and descriptive questions.

\textbf{Data Statistics.}
The {\dataset} dataset consists of five types of question outputs: binary yes/no questions (32\%), point-in-time questions (14\%), duration questions (4\%), spatial position questions (34\%), and descriptive text questions (16\%).
Figure~\ref{fig:dataset} depicts the distribution over time of the videos and examples of questions.
These questions focused on spatial understanding, object detection, sign reading, dynamic event understanding, and contextual reasoning.

\definecolor{Gray}{gray}{0.9}

\newcolumntype{a}{>{\columncolor{Gray}}c}

\begin{table*}[t]
\centering
\vspace{5px}

\setlength{\aboverulesep}{0pt}
\setlength{\belowrulesep}{0pt}
\setlength{\extrarowheight}{.75ex}
\setlength{\tabcolsep}{4pt}

\begin{tabular}{@{}ca
ccc
aaa 
ccc@{}}

\toprule
\textbf{Method} &     \textbf{LLMs} & \multicolumn{3}{c}{\begin{tabular}[c]{@{}c@{}}\textbf{Descriptive Question} \textbf{Accuracy $\uparrow$}\end{tabular}} & \multicolumn{3}{c}{\cellcolor{Gray}\begin{tabular}[c]{@{}c@{}}\textbf{Positional Error (m) $\downarrow$}\end{tabular}} & \multicolumn{3}{c}{\begin{tabular}[c]{@{}c@{}}\textbf{Temporal Error (s) $\downarrow$}\end{tabular}}  \\ \midrule

 & & \multicolumn{1}{c}{Short} & \multicolumn{1}{c}{Medium} & \multicolumn{1}{c}{Long} & \multicolumn{1}{c}{\cellcolor{Gray}Short} & \multicolumn{1}{c}{\cellcolor{Gray}Medium} & \multicolumn{1}{c}{\cellcolor{Gray}Long} & \multicolumn{1}{c}{Short} & \multicolumn{1}{c}{Medium} & \multicolumn{1}{c}{Long} \\ 

\midrule
        Ours & GPT4o & \textbf{0.62\scriptsize{$\pm$0.5}} & {\textit{0.58\scriptsize{$\pm$0.5}}} & \textbf{0.65\scriptsize{$\pm$0.5}} & \textbf{5.1\scriptsize{$\pm$11.9}} & {\textbf{27.5\scriptsize{$\pm$26.8}}} & \textbf{46.25\scriptsize{$\pm$59.6}} & \textbf{0.3\scriptsize{$\pm$0.1}} & \textbf{1.8\scriptsize{$\pm$2.0}} & \textbf{3.6\scriptsize{$\pm$5.9}} \\
                
& Codestral & 0.25\scriptsize{$\pm$0.4} & 0.24\scriptsize{$\pm$0.4} & 0.11\scriptsize{$\pm$0.3} & 151.3\scriptsize{$\pm$109.7} & 189.0\scriptsize{$\pm$109.6} & 212.4\scriptsize{$\pm$121.3} & 4.8\scriptsize{$\pm$5.6} & 8.4\scriptsize{$\pm$6.8} & 14.8\scriptsize{$\pm$7.5} \\
& Command-R & 0.36\scriptsize{$\pm$0.5} & 0.32\scriptsize{$\pm$0.5} & 0.14\scriptsize{$\pm$0.3} & 158.7\scriptsize{$\pm$129.6} & 172.2\scriptsize{$\pm$119.4} & 188.7\scriptsize{$\pm$107.1} & 4.5\scriptsize{$\pm$17.3} & 14.3\scriptsize{$\pm$6.7} & 15.3\scriptsize{$\pm$11.7} \\
& Llama3.1:8b & 0.31\scriptsize{$\pm$0.5} & 0.33\scriptsize{$\pm$0.5} & 0.21\scriptsize{$\pm$0.4} & 159.9\scriptsize{$\pm$123.2} & 151.2\scriptsize{$\pm$121.1} & 165.3\scriptsize{$\pm$115.1} & 9.5\scriptsize{$\pm$27.5} & 7.9\scriptsize{$\pm$16.3} & 18.7\scriptsize{$\pm$10.8} \\

\midrule

\multirow{1}{*}{LLM with Caption} 
        & GPT4o & \textit{0.57\scriptsize{$\pm$0.5}} & \textbf{0.66\scriptsize{$\pm$0.5}} & {\textit{0.55\scriptsize{$\pm$0.5}}} & {\textbf{5.1\scriptsize{$\pm$8.2}}} & \textit{33.3\scriptsize{$\pm$47.3}} & {\textit{56.0\scriptsize{$\pm$61.7}}} & {\textit{0.5\scriptsize{$\pm$0.5}}} & {\textit{1.9\scriptsize{$\pm$2.2}}} & {\textit{8.0\scriptsize{$\pm$6.7}}} \\
\midrule
Multi-Frame VLM 
        & GPT4o    & \textit{0.55\scriptsize{$\pm$0.5}} & \xmark & \xmark & \textit{7.5\scriptsize{$\pm$11.4}} & \xmark & \xmark & 0.5\scriptsize{$\pm$2.2} & \xmark & \xmark  \\ 
\bottomrule

\end{tabular}
\caption{\textbf{Results.} 
We compare {\model} to an approach that processes all captions at once and another that processes all frames at once. 
We find that GPT4o-based approaches perform the best, and  that {\model} outperforms the LLM-based method and remains competitive to the VLM-based approach on the Short videos.
The Medium and Long videos are too long for the VLM to process, and thus is marked with an \xmark.}
\label{tab:correctness}
\end{table*}

\section{Experimental Setup}
We use {\dataset} to evaluate the ability of {\model} and other LLM-based approaches.

\textbf{Methods.}
{\model} uses a retrieval module to aggregate relevant parts of the long-horizon history.
We show the ability of {\model} with a closed-source LLM (GPT-4o), various open-source LLMs (Codestral~\cite{codestral_model_card}, Command-R~\cite{command_r_model_card}), and a smaller 8 billion parameter Llama3.1~\cite{dubey2024llama} model. 
{\model} uses up to 3 retrieval steps to construct $R$. 
We compare these models to using GPT-4o with all the captions provided at once and a version using frames sampled at 2 FPS from the video itself.
For captioning, we use the VILA1.5-13b over 3 seconds of video, leading to 2 FPS.

\textbf{Metrics.}
The {\dataset} dataset consists of four types of answers, for which we compute different metrics.
To unify each of these types of metrics into one metric and reduce the impact of outliers, we threshold the temporal and spatial metrics to determine whether an instance is correct or not to create an \textbf{Overall Correctness} metric.

\begin{itemize}
    \item Spatial questions output (x,y,z) coordinates, from which we compute an L2 distance. We define a spatial question to be correct if it is within 15 meters of the goal.
    \item Temporal point-in-time and duration questions produce answers such as ``15 minutes'', for which we compute L1 temporal error. We define a temporal question to be correct if it is within 2 minutes of the goal.
    \item Descriptive questions produce either yes/no or textual answers, for which we compute a binary accuracy. This accuracy also determines correctness.
\end{itemize}

To make evaluation faster, text answers are evaluated by an LLM to be correct or not, similar to other work~\cite{majumdar2024openeqa}.

All {\model} experiments are run over three seeds while the baseline results are over one seed due to cost. 
Since seeds are not as reproducible, we micro-average the results across all seeds.
The variance is high due to the differences in difficulty between questions.

\begin{figure}[b]
\centering
\includegraphics[width=1\linewidth]{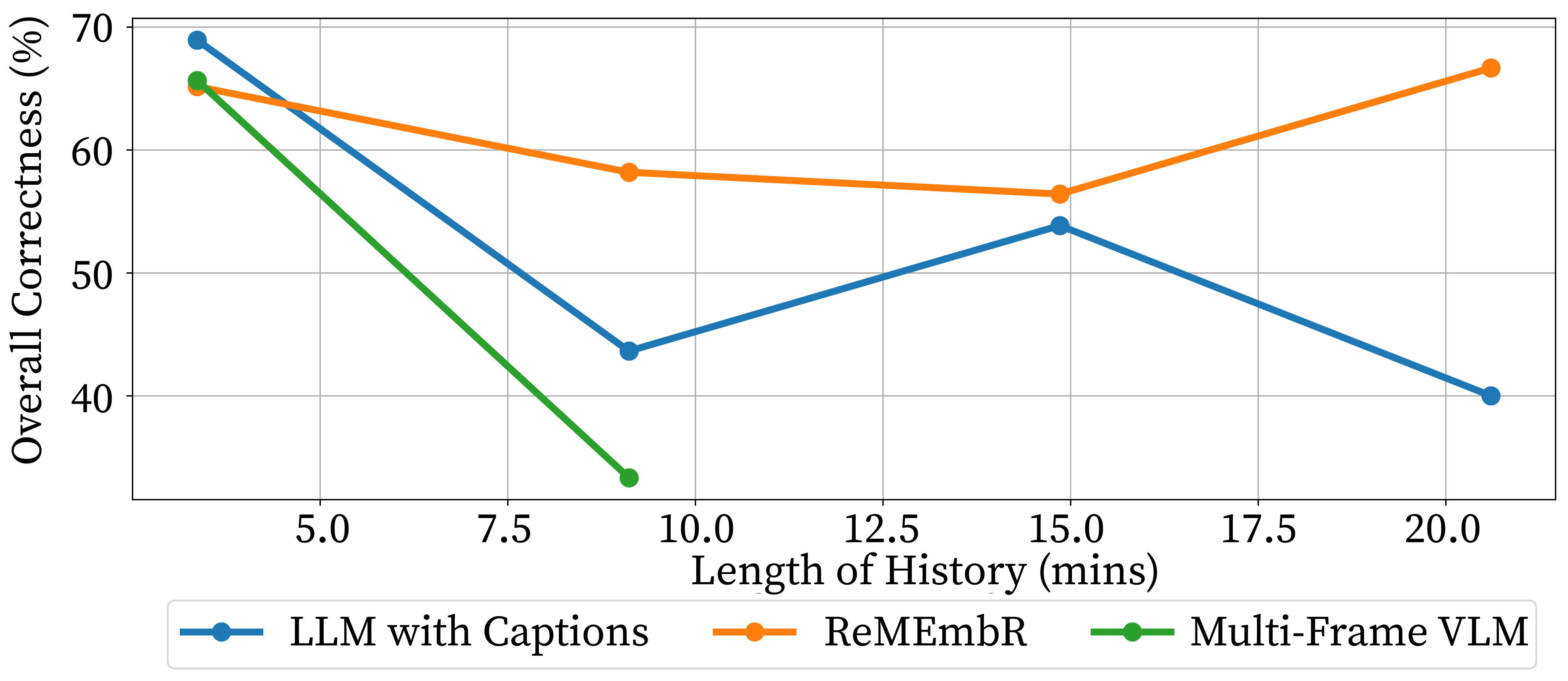}
\caption[]{\textbf{Overall correctness over time.} We discretize time into 4 bins and average overall correctness scores in each. Note that although the Medium category in Table~\ref{tab:correctness} is incomplete, some test instances did complete. We find that {\model} is more correct as the amount of time increases.}
\label{fig:correctness_time}
\end{figure}

\section{Results}

\textbf{{\model} performs strongly given a long-horizon memory at a lower latency.}
As shown in the results in Table~\ref{tab:correctness}, {\model} improves performance on long-horizon tasks compared to traditional LLM methods.
For long-duration videos, {\model} using GPT4o achieves better descriptive question accuracy, positional error, and temporal error compared to the LLM with captions and Multi-Frame VLM baselines.
{\model} performed similarly to the VLM for short category; however, the VLM is unable to process the long videos and most of the medium length videos.

\textbf{{\model} scales to longer videos with higher overall corectness.}
Figure~\ref{fig:correctness_time} shows the overall correctness over time. Although {\model} does not have the highest performance for short videos compared to the VLM and LLM with captions, {\model} is able to maintain a higher overall correctness score as the video length scales to be longer.

\textbf{{\model} performs with low latency.}
We found that for a 21.5 minute video, {\model} takes approximately $25$ seconds per question, while the VLM took around $90$ seconds per question for a shorter 5.5 minute video.
In fact, since {\model} only calls retrieval functions, the amount of time to answer a question remains relatively static regardless of the video duration.
Despite their lower performance, we also note that Command-R and Codestral running on a local desktop takes around $40$ seconds, while the smaller Llama3.1-8b takes around $15$ seconds.


\textbf{Open-source LLMs perform worse than GPT-4o.}
As shown in Table~\ref{tab:correctness}, we found that LLMs trained specifically for code or function calling work well for generating queries.
However, our results imply that these LLMs struggle largely with arithmetic reasoning required for answering temporal and spatial questions, leading to lower performance.


\textbf{Longer caption lengths hurt performance.}
We captioned with VILA1.5-13b during memory building by passing the model 6 frames for every 3 seconds of accumulated video, effectively operating at 2 FPS.
We chose 6 frames as this is the max number of frames VILA can process.
To evaluate the effect of frame rate, we also tested a lower rate of 6 frames every 12 seconds, or 0.5 FPS.
We observed that captioning at this reduced frame rate led to a drop in performance, likely due to information loss from the coarser sampling.

\textbf{Different sizes of captioning models slightly reduces performance.}
As shown in Table~\ref{tab:ablations}, using the 13b captioning model performs slightly better than smaller 8b and 3b models with respect to overall correctness.
The minimal performance loss for using the 3b model is important as smaller models have a higher throughput when deployed on a robot.

\textbf{Iterative function calls are required for good performance.}
{\model} uses up to three iterations to find the answer. 
We found that with only one iteration, which is similar to traditional retrieval-augmented generation, overall correctness decreases.
This is likely due to some questions requiring multi-step reasoning, or if the first retrieval did not provide relevant information, {\model} can try again.
\definecolor{Gray}{gray}{0.9}

\newcolumntype{d}{>{\columncolor{Gray}}l}


\begin{table}[t]
\centering
\vspace{5px}

\setlength{\aboverulesep}{0pt}
\setlength{\belowrulesep}{0pt}
\setlength{\extrarowheight}{.5ex}
\begin{tabular}{dlll@{}}

\toprule
\multicolumn{1}{a}{\textbf{LLMs}} & \multicolumn{3}{c}{\begin{tabular}[c]{@{}c@{}}\textbf{Overall} \textbf{Correctness $\uparrow$}\end{tabular}} \\ \midrule

 & \multicolumn{1}{c}{Short} & \multicolumn{1}{c}{Medium} & \multicolumn{1}{c}{Long} \\ 

\midrule
ReMEmbR     & \textbf{0.72\scriptsize{$\pm$0.5}} & \textbf{0.56\scriptsize{$\pm$0.5}} & \textbf{0.61\scriptsize{$\pm$0.5}} \\
\midrule
\quad- 1 call only & 0.67\scriptsize{$\pm$0.5} & 0.48\scriptsize{$\pm$0.4} & 0.50\scriptsize{$\pm$0.5} \\
\midrule
\quad - 12-sec captions & 0.54\scriptsize{$\pm$0.5} & 0.50\scriptsize{$\pm$0.5} & 0.38\scriptsize{$\pm$0.5} \\ 

\midrule
\quad - Llama-VILA1.5-8b & 0.58 \scriptsize{$\pm$0.5} & 0.52 \scriptsize{$\pm$0.5} & 0.54\scriptsize{$\pm$0.5} \\
\quad - VILA1.5-3b & 0.60\scriptsize{$\pm$0.5} & 0.58\scriptsize{$\pm$0.5} & 0.50\scriptsize{$\pm$0.5} \\ 

\bottomrule

\end{tabular}
\caption{\textbf{Ablations.} We provide various abaltions of different components of {\model}. We find that the iterative querying process, 3-second captions, and the size of the captioning model are important components to making {\model} work.}
\label{tab:ablations}
\end{table}



\section{Real World Deployment}

Though {\dataset} is useful for prototyping and validating new methods, it is important to deploy such methods on robots.
In this section, we demonstrate that {\model} can also be deployed in real time on a robot in the real world.

\textbf{Robot Deployment.} We deploy {\model} on a Nova Carter robot~\cite{SegwayNovaCarter}.
We run the memory building phase on Jetson Orin 32GB, and use GPT-4o as the LLM backend for the {\model} agent.
We run a quantized version of VILA-3b to aggregate captions over time. 
We use ROS2's Nav2 stack with AMCL for computing localization over a pre-mapped metric map.
We run a Whisper automatic speech recognition model~\cite{radford2023robust} that was optimized for a Jetson to enable interaction with {\model}.
VILA-3b, Whisper, the Nav2 stack with 3D LiDAR, and the vector database querying runs on-device. 
In the code release, we will provide various LLM backends such as cloud-based LLMs like NVIDIA NIM APIs~\cite{nv_nim} or OpenAI APIs, local large LLMs like Command-R that can run on a local desktop, and smaller function-calling LLMs that can run on-device. 
We hope that our code release can enable researchers to build and query long-horizon robot histories across arbitrary embodiments. 

\begin{figure}[t]
\centering
\vspace{5px}
\includegraphics[width=1\columnwidth]{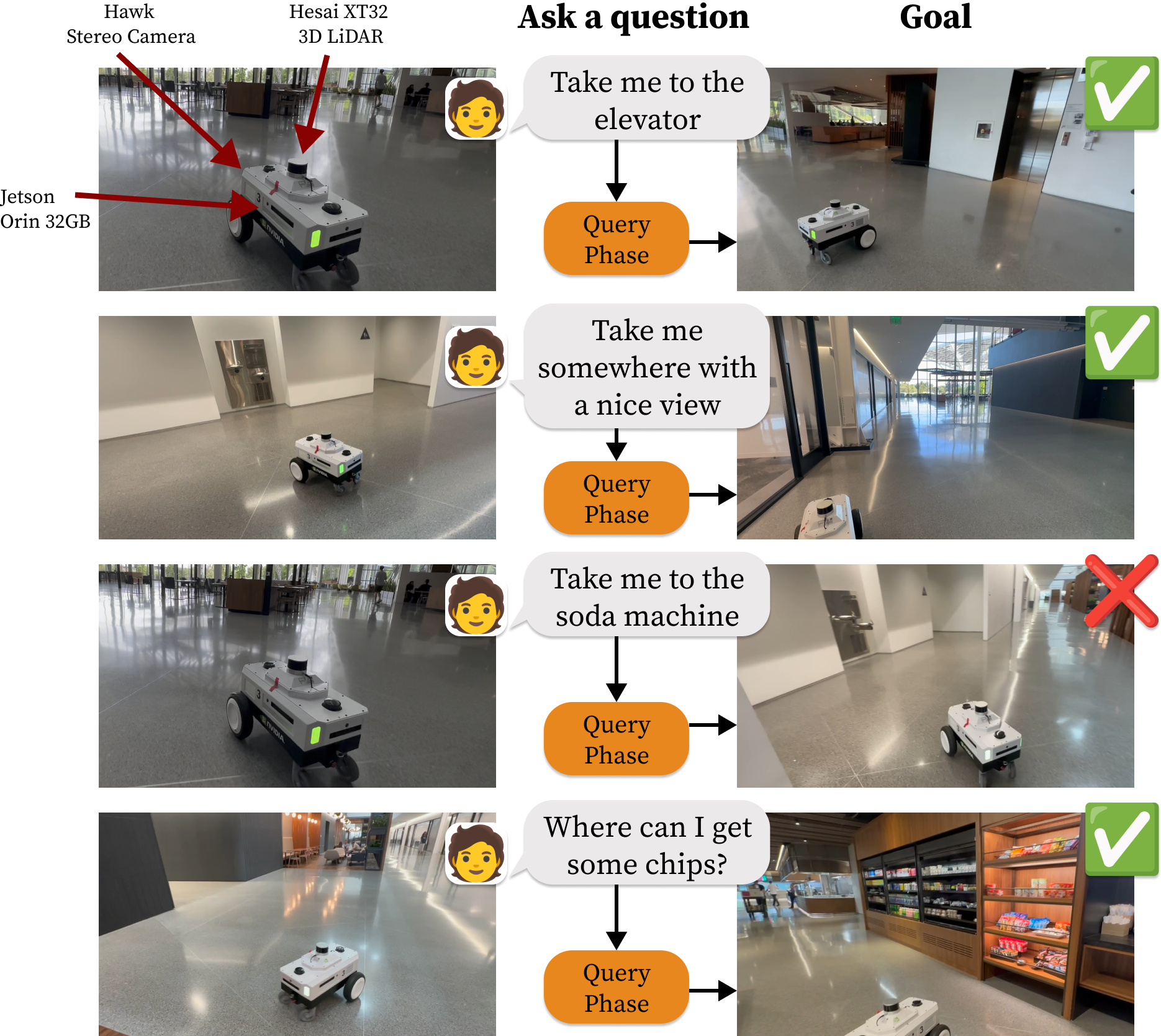}
\caption[]{
\textbf{Robot deployment.} 
We deploy {\model} on a Nova Carter robot.
We run the memory building phase for 25 minutes, and then begin to ask navigation-centric questions
The robot successfully handles various instructions, including those with more ambiguous instructions such as going to somewhere with a nice view. 
However, we found that {\model} often confuses some objects such as soda machines and water fountains, leading to incorrect goals.
}
\label{fig:deployment}
\end{figure}


\textbf{Qualitative results.}
We deployed the system in a large office space by first building a memory by driving the robot around for 25 minutes.
Then we began querying the robot with various navigation-centric questions.
We found that our robot was able to execute tasks such as ``Where can I get some chips" where the robot took the user to a cafeteria shelf that contained chips.
In contrast to searching for specific objects, we also found that our system can guide users to more general areas such as food courts if asked about food or drinks.
Our system can also handle more vague questions.
We asked the robot to ``Take me somewhere with a nice view", and observed the function calls looking for tall glass windows, plants, and open spaces. 
Then the robot navigated to a lobby with large glass windows and greenery.
We also found that for questions such as ``Take me to the soda machine", the robot would go to a water fountain, as it was captioned as a ``silver machine". 
This is likely an artifact of using a quantized 3B captioning model that was unable to caption the water fountain properly.

\section{Conclusion}
In this work, we introduced {\model}, a system designed to address the challenge of long-horizon video question answering for robots.
By decomposing the task into a memory building phase using a VLM and a vector database then a querying phase with an LLM-agent, {\model} efficiently handles the extensive histories that robots accumulate over time.
This approach makes it feasible for robots to leverage long-term memory in dynamic and complex environments.

\textbf{Limitations and Future Work.}
While {\dataset} ensures a unique answer for each question, real-world deployments often involve situations where multiple potential answers could be valid, which would require more focus on contextual reasoning.
Additionally, our memory-building approach relies solely on video captioning.
However, real-world environments contain rich spatial information such as room numbers, equipment labels, and other details that could be manually annotated.
Semantic maps, scene graphs, and queryable scene representations can also provide useful spatial information.
We hope to integrate other kinds of memory as function calls so that the {\model} agent can reason spatially and contextually across a broader range of information. 
A limitation of our approach is that it constantly adds potentially repetitive information into the vector database which would dilute useful information over time.
We believe that efficient memory aggregation of pertinent information is an interesting area of future research.




\bibliographystyle{IEEEtran}
\bibliography{refs}

\end{document}